\documentclass{article}

\usepackage{PRIMEarxiv}

\usepackage[utf8]{inputenc} 
\usepackage[T1]{fontenc}    
\usepackage{hyperref}       
\usepackage{url}            
\usepackage{booktabs}       
\usepackage{amsfonts}       
\usepackage{nicefrac}       
\usepackage{microtype}      
\usepackage{lipsum}
\usepackage{fancyhdr}       
\usepackage{graphicx}       
\usepackage{amsmath}
\usepackage{tabularx}
\usepackage{array}
\usepackage{bbding}
\usepackage{amssymb}
\usepackage{multicol}
\usepackage{multirow}
\usepackage{mwe}
\usepackage{pdfpages}
\usepackage{float}
\usepackage{amssymb}
\usepackage{algorithm}
\usepackage{algorithmicx}
\usepackage{algpseudocode}
\usepackage{subfigure}
\usepackage{color}
\usepackage{xcolor}
\graphicspath{{media/}}     

\author{
    Xueru Wen \\
    College of Computer Science and Technology \\
    Jilin University \\
    Changchun\\
    \texttt{wenxr2119@mails.jlu.edu.cn} \\
\And
    Changjiang Zhou \\
    College of Computer Science and Technology \\
    Jilin University \\
    Changchun\\
\And
    Haotian Tang \\
    College of Computer Science and Technology \\
    Jilin University \\
    Changchun\\
\And
    Luguang Liang \\
    College of Computer Science and Technology \\
    Jilin University \\
    Changchun\\
\And
    Yu Jiang \\
    Key Laboratory of Symbolic Computation and Knowledge Engineering of Ministry of Education \\
    Jilin University \\
    \texttt{jiangyu2011@jlu.edu.cn} \\
\And
    Hong Qi \\
    Key Laboratory of Symbolic Computation and Knowledge Engineering of Ministry of Education \\
    Jilin University \\
}

\title{End-to-End Entity Detection with Proposer and Regressor}

\begin{document}
\maketitle

\begin{abstract}
Named entity recognition is a traditional task in natural language processing. 
In particular, nested entity recognition receives extensive attention for the widespread existence of the nesting scenario.
The latest research migrates the well-established paradigm of set prediction in object detection to cope with entity nesting.
However, the manual creation of query vectors, which fail to adapt to the rich semantic information in the context, limits these approaches.    
An end-to-end entity detection approach with proposer and regressor is presented in this paper to tackle the issues.
First, the proposer utilizes the feature pyramid network to generate high-quality entity proposals.
Then, the regressor refines the proposals for generating the final prediction.
The model adopts encoder-only architecture and thus obtains the advantages of the richness of query semantics, high precision of entity localization, and easiness of model training.
Moreover, we introduce the novel spatially modulated attention and progressive refinement for further improvement.
Extensive experiments demonstrate that our model achieves advanced performance in flat and nested NER, achieving a new state-of-the-art F1 score of 80.74 on the GENIA dataset and 72.38 on the WeiboNER dataset.
\end{abstract}


\section{Introduction}
\begin{figure}[H]
    \centering
    \includegraphics[width=\textwidth]{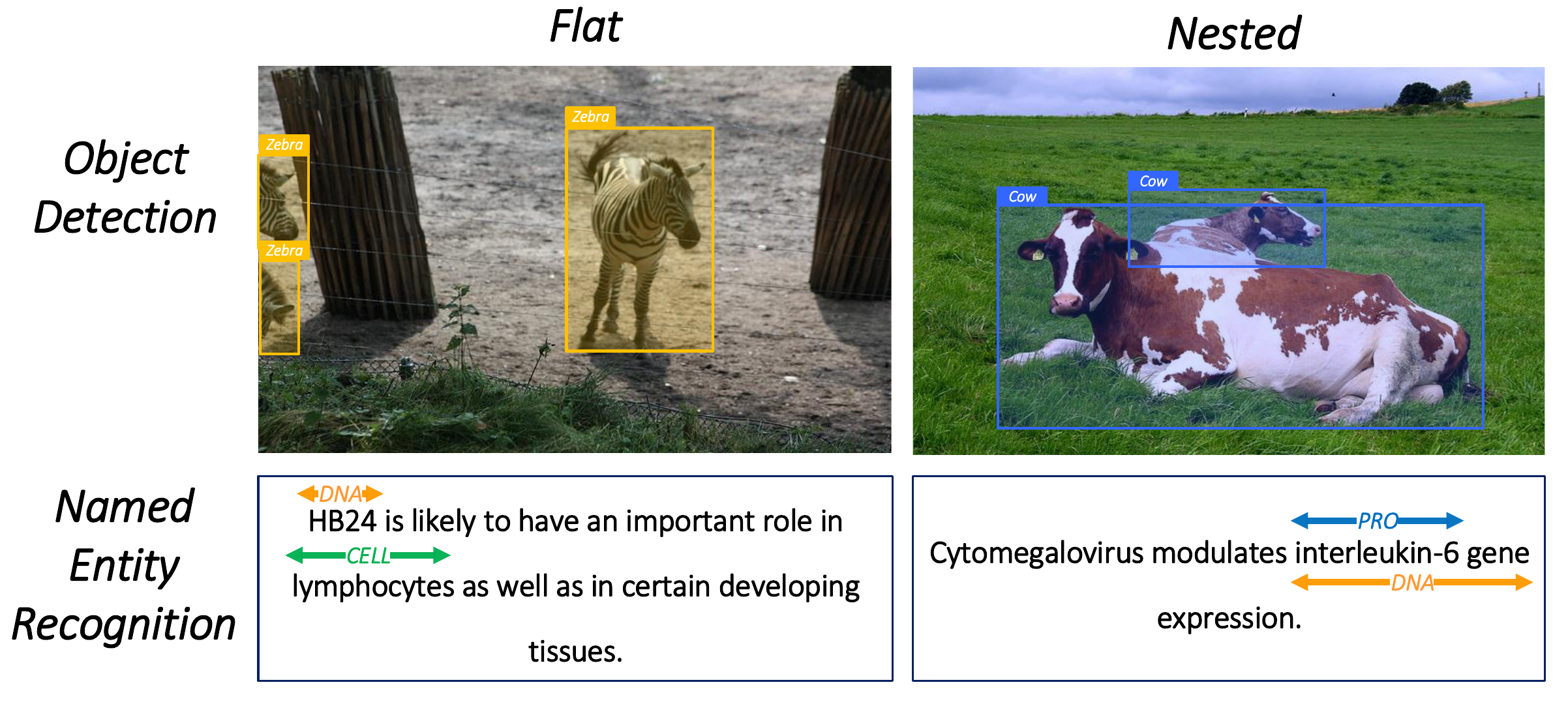}
    \caption{Examples for object detection and named entity recognition under flat and nested circumstances. Examples are obtained from GENIA \cite{10.5555/1289189.1289260} and COCO2017 \cite{10.1007/978-3-319-10602-1_48}.}
    \label{example}
\end{figure}

Named entity recognition identifying text spans of specific entity categories is a fundamental task in natural language processing.
It has played a crucial role in many downstream tasks such as relation extraction \cite{wang_entity_2022}, information retrieval \cite{10.1145/3331184.3331333} and entity linking \cite{CHEN201812}.
The model \cite{liu_tfm_2022, yan_named_2021} based on sequence labeling has achieved great success in this task.
Even though mature and efficient as these models are, they fail to handle the nested entities that are a non-negligible scenario in the real-world language environment.
Some recent studies \cite{shen-etal-2021-locate} have noted the formal consistency of Object detection with NER tasks.
Figure \ref{example} shows the instances where the entities overlap with each other and the detection boxes intersect with each other.

A few previous works have designed proprietary structures to deal with the nested entities, such as the constituency graph \cite{finkel-manning-2009-nested} and hypergraph \cite{HUANG2021200}.
Other works \cite{alex-etal-2007-recognising, fisher-vlachos-2019-merge} capture entities through the layered model containing multiple recognition layers.
Despite the success achieved by these approaches, they inevitably necessitate the deployment of sophisticated transformations and costly decoding processes, introducing extra errors compared to the end-to-end manner.

Seq2Seq methods \cite{ju-etal-2018-neural} can address various kinds of NER subtasks in a unified form.
However, these methods have difficulties defining the order of the outputs due to the natural conflict between sets and sequences.
This trait limits the performance of the model.
The span-based approaches \cite{xu-etal-2017-local, sohrab-miwa-2018-deep}, which identify entities by enumerating all candidate spans in a sentence and classifying them, also receive lots of attention.
Although enumeration can be theoretically perfect, the high computational complexity still burdens these methods.
Second, these methods mainly focus on learning span representations without the supervision of entity boundaries \cite{Tan2020BoundaryEN}. 
Further, enumerating all subsequences from a sentence generate many negative samples, which reduces the recall rate.
Some recent work, including set prediction networks, has attempted to address these defects.

The latest works \cite{shen-etal-2022-piqn} treat information extraction as a reading comprehension task, extracting entities and relations through manually constructed queries.
The set prediction network \cite{DianboSui2020JointEA} is introduced to the entity and relation extraction.
Because the techniques accommodate the unordered character of the prediction target, these methods achieve great success.
However, most of them still confront problems caused by query vectors.
The random initialization of the query vector leads to the lack of sufficient semantic information and difficulty in learning the proper attention pattern.

This paper presents the end-to-end entity detection network, which predicts the entities in a single run and thus is no longer affected by prediction order.
The proposed model transforms the NER task into a set prediction problem.
First, we utilize the feature pyramid network to build the proposer, which generates high-quality entity proposals with rich semantical query vectors, high-overlapping spans, and category logarithms.
High-quality proposals significantly alleviate the difficulty of training.
Then, the encoder-only regressor constructed by the iterative transformer employs the regression procedure on the entity proposals.
In contrast to some span-based methods discarding the partially match proposals,  the regressor adjusts these proposals to improve model performance.
The prediction head computes probability distributions for each entity proposal to identify the entities.
In the training phase, we dynamically assign prediction targets to each proposal.

Moreover, we introduce the novel spatially modulated attention in this paper.
It guides the model to learn more reasonable attention patterns and enhances the sparsity of the attention map by making full use of the spatial prior knowledge, which improves the model's performance.
We also correct the entity proposal at every layer of the regressor network, called the progressive refinement.
This strategy increases the precision of the model and facilitates gradient backpropagation.

Our contribution can be summarized as follows:
\begin{itemize}
    \item We design the proposer constructed by the feature pyramid to incorporate multi-scale features and initialize high-quality proposals with high-overlapping spans and strongly correlated queries.
    Compared to previous works which randomly initialize query vectors, the proposer network may greatly reduce training difficulty.
    \item  We deploy the encoder-only framework in the regressor, which evades the handcraft construction of query vectors and hardships in learning appropriate query representations and thus notably eases the difficulties of convergence.
    The iterative refinement strategy is further utilized in the regressor to improve the precision and promote gradient backpropagation.
    \item We introduce the novel spatially modulated attention mechanisms that help learn the proper attention pattern. 
    The spatially modulated attention mechanism dramatically improves the model's performance by integrating the spatially prior knowledge to increase attention sparsity.
\end{itemize}

\section{Related Work} 
In this section, we will review some related works about Named entity recognition and set prediction.
We will analyze different methodologies for the NER task and the trend of set prediction algorithm development.

\subsection{Named Entity Recognition}
Since traditional NER methods with sequence labeling \cite{liu_ltp_2022, wang_learning_2019} have been well studied, many works \cite{AlejandroMetkeJimenez2016ConceptIA} have been devoted to extending sequence tagging methods to nested NER.
One of the most explored methods is the layered approach \cite{wang-etal-2020-pyramid}.
Other works deploy proprietary structures to handle the nested entities, such as the hypergraph \cite{HUANG2021200}.
Although these methods have achieved advanced performance, they are still not flexible enough due to the need for manually designed labeling schemes.
In contrast, the end-to-end framework proposed in this paper avoids this disadvantage and thus facilitates the implementation and migration of the method.

Seq2Seq approach \cite{yan-etal-2021-unified-generative} unifies different forms of nested entity problems into sequence generation problems.
Even though this strategy avoids the complicated annotation methods, the sensitivity of the decoding order and beam search algorithm poses a barrier to boosting the model performance.
The end-to-end model in this paper incorporates the set prediction algorithm in order to overcome the difficulties confronting the Seq2Seq model. 

Span-based approaches \cite{ChuanqiTan2020BoundaryEN, LI202126}, which classify candidate spans to identify the entities, also draw broad interest.
The span-based method formally resembles the object detection task in computer vision. 
Based on the long-standing idea \cite{Manning2016CONNECTINGIA} of associating the image with natural language, some \cite{shen-etal-2021-locate} proposes a two-stage identifier that fully exploits partially matched entity proposals and introduces the regression procedure.
Despite the instructiveness of the insight to migrate proven methods in computer vision to NER tasks, the error propagation due to two-stage models and the proper way for boundary regression in languages remain issues to address.
In this paper, we comprise the propose and regression stage into an end-to-end framework and refine the entity proposal based on probability distribution to fit the language background of the NER task.

\subsection{Set Prediction}
Several recent pieces of research \cite{DianboSui2020JointEA, ijcai2021-542} have deployed the set prediction network in information extraction tasks and proved its effectiveness.
These works can be seen as variants of \textit{DETR} \cite{10.1007/978-3-030-58452-8_13} which is proposed for object detection and use the transformer decoder to update the manually created query vector for the generation of detection boxes and corresponding categories.

The models based on set prediction networks, especially \textit{DETR}, have been extensively studied.
Slow convergence due to the random initialization of the object query is the fundamental obstacle of \textit{DETR}.
A two-stage model \cite{ZhiqingSun2020RethinkingTS} with the feature pyramid \cite{TsungYiLin2016FeaturePN} which generates high-quality queries and introduces multi-scale features is proposed to settle the convergence problem.
This paper also discusses the necessity for cross-attention and suggests that an encoder-only network can achieve equally satisfactory results.
Spatially modulated co-attention \cite{9709993} integrating spatially prior knowledge is also introduced to ease the problem.
This work increases the sparsity of attention through a priori knowledge for the purpose of accelerating training.
The thought-provoking deformable attention
\cite{XizhouZhu2021DeformableDD} is presented, which shows the possibility of models learning the spatial structure of the attention.
It also improves the model performance by iteratively refining the detection box.

A considerable amount of improvement work has been proposed to solve the convergence of set prediction problems.
Previous work \cite{ijcai2021-542} employing set prediction requires delicate selection for the number of queries to reach a promising speed of model convergence.
Inspired by the analysis \cite{ZhiqingSun2020RethinkingTS} done previously, we propose the end-to-end framework to settle the problem.

\section{Method}
In this section, we are going to detail our method.
The general framework of our model is shown in Figure \ref{Framework}, which is constructed in the following parts:

\begin{figure}
    \centering
    \includegraphics[width=\textwidth]{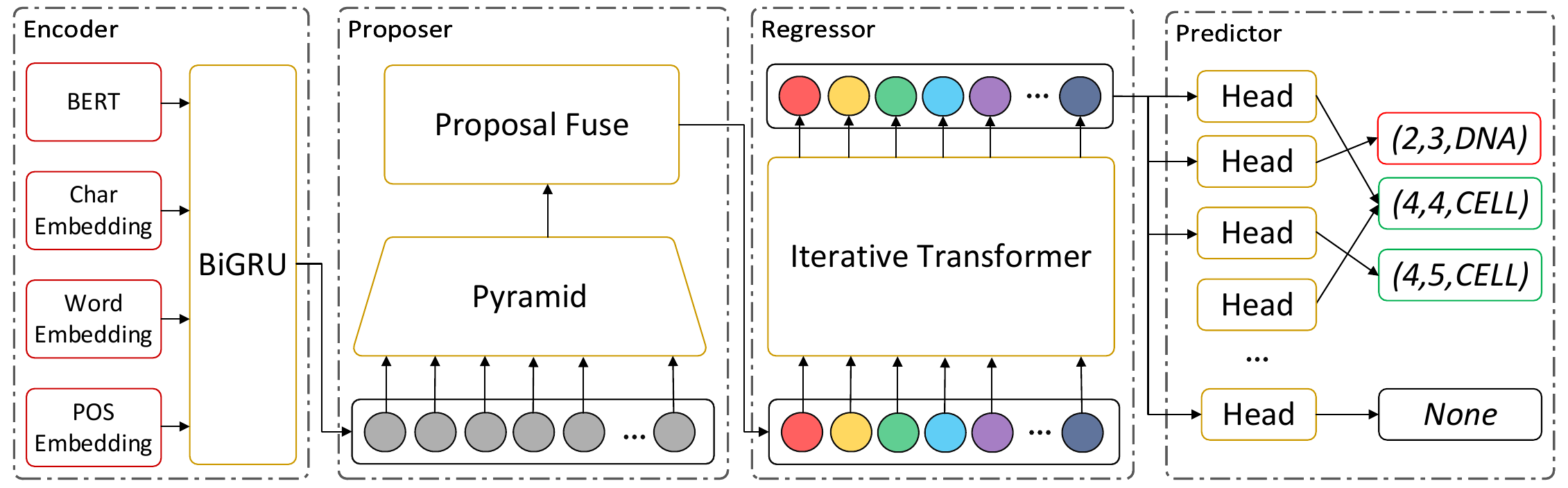}
    \caption{Architecture of our proposed model for end-to-end entity detection.}     
    \label{Framework}
\end{figure}

\begin{itemize}
    \item{\textbf{Sentence Encoder}}
    We utilize hybrid embedding to encode the sentence. 
    The generated embeddings are then fused by the \textit{BiGRU} \cite{cho2014learning} to produce the final multi-granularity representation of the sentence.
    \item{\textbf{Proposer}}
    We build up the feature pyramid through the stack of \textit{BiGRU} and \textit{CNN} \cite{kim-2014-convolutional} to constitute the proposer network.
    The proposer exploits the multi-scale features to initialize the entity proposals.
    \item{\textbf{Regressor}}
    We design the regressor that refines the proposals progressively to locate and classify spans more accurately.
    The regressor is built by stacking update layers constructed by the spatially modulated attention mechanism.
    \item{\textbf{Prediction Head}}
    The prediction head outputs the span location probability distribution based on the refined proposals.
    The distribution will be combined with probabilities generated by the category logarithms to compute the joint probability distribution, from which can obtain the eventual prediction results.
\end{itemize}

\subsection{Sentence Encoder}
The goal of this component is to transform the original sentence into dense hidden representations.
With the inputted sentence $\mathcal{S}=\left[w_1,w_2,...w_L\right]$, we represent the $i$-th token $w_i$ with the concatenation of multi-granularity embeddings as follows:
\begin{equation}
    h^{\prime}_i = \left[h^{char}_i;h^{bert}_i;h^{word}_i;h^{pos}_i \right] 
\end{equation}

The embedding at character level $h^{char}$ is generated by fusing each character's embedding $h^c$ through recurrent neural networks and average pooling them as follows:
\begin{equation}
h^{char}_i=\textit{Pool}\left(\textit{BiGRU}\left(\left[h^c_1,h^c_2,... ,h^c_{D}\right]\right)\right)
\end{equation}
where $D$ is the number of characters constituting the token.
The character-level embedding can help the model cope better with out-of-vocabulary words.

$h^{bert}$ stands for the representation generated by the pre-trained language model \textit{BERT} \cite{devlin2019bert}.
We follow \cite{wang-etal-2020-hit} to obtain the contextualized representation by encoding the sentence with the surrounding tokens.
The \textit{BERT} separates the tokens into subtokens by Wordpiece partitioning \cite{wu2016google}.
The representation of subtokens is average pooled to create the contextualized embedding as follows:
\begin{equation}
h^{bert}_i=\textit{Pool}\left(\left[h^b_1,h^b_2,... ,h^b_O\right]\right)
\end{equation}
where $O$ is the number of subtokens forming the token.
The pre-trained model can aid in the generation of more contextually relevant text representations.

As for the embedding of word-level $h^{word}_i$, we exploit pre-trained word vectors including \textit{Glove}  \cite{pennington-etal-2014-glove}.
To introduce the semantic message of part-of-speech, we embed each token's POS tag as $h^{pos}_i$.

The multi-granularity embeddings are then fed into the \textit{BiGRU} network to produce the hybrid embedding for the final representation $H$ of the sentence as follows:
\begin{equation}
    \begin{aligned}
        H^{\prime} &= \left[h^{\prime}_1,h^{\prime}_2,...,h^{\prime}_L\right]\\
        H &= W \textit{BiGRU}\left(H^{\prime}\right) + b\\
            &= \left[h_1,h_2,...,h_L\right]\\
    \end{aligned}
\end{equation}

\subsection{Proposer}
\begin{figure}
    \centering
    \includegraphics[width=\textwidth]{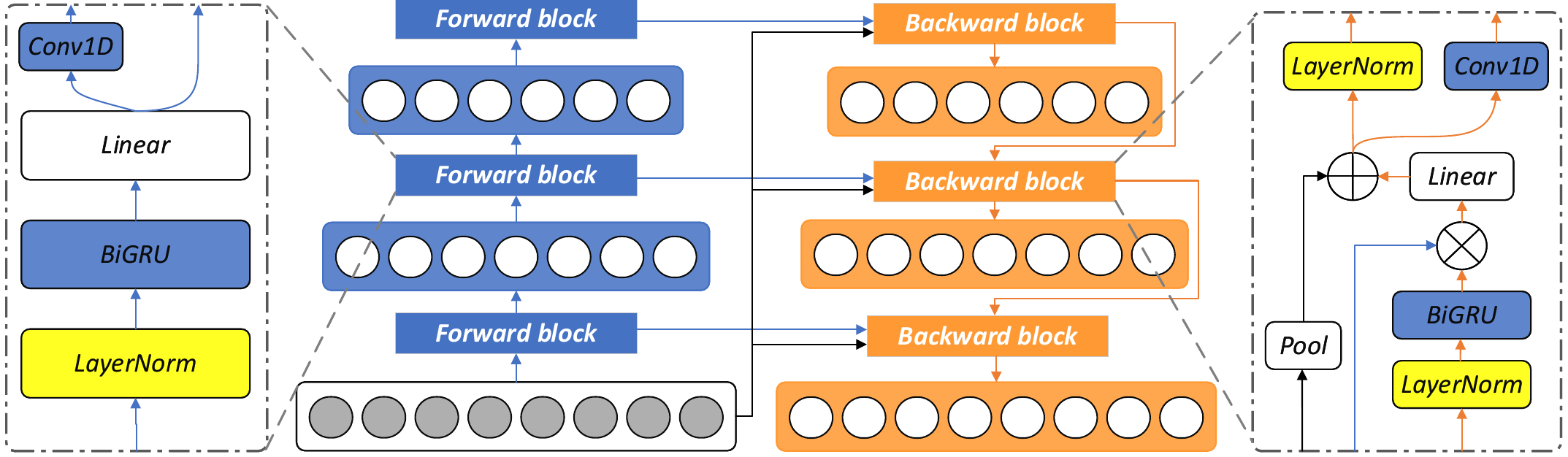}
    \caption{Data flow of feature pyramid and detailed structure of blocks.}     
    \label{pyramid}
\end{figure}

We tailor the pyramid network \cite{wang-etal-2020-pyramid} to build up our proposer, which is able to integrate features of multi-scales and reasonably initialize the proposals.
Figure \ref{pyramid} illustrates the structure of the feature pyramid.
The constitution of the pyramid is performed in a bottom-to-up and up-to-bottom manner.
The bidirectional construction procedures allow better message communication between layers.
We selectively merge the features at different layers to yield initial proposals.
This process is implemented in a similar way to the attention mechanism.

\subsubsection{Forward Block}
The feature pyramid is first built from the bottom to up. 
It consists of $L$ layers, each with two main components, a \textit{BiGRU} and a \textit{CNN} of kernel sizes $k$. 
At layer $l$, the \textit{BiGRU} models the interconnections of spans of the same size.
The \textit{CNN} aggregates $k$ neighboring hidden states, which are then passed into higher layers. 
Apparently, each feature vector represents a span of $v_l$ original tokens and $v_l$ can be calculated as:
\begin{equation}
    v_l = 1 - l + \sum_{i=1}^{l}k_i
\end{equation}

One may note that the pyramid structure provides inherent induction: the higher the number of layers, the shorter the input sequence, with higher levels of feature vectors representing the long entities and lower levels representing the short entities.
Moreover, since the input scales of the layers are diverse, we apply \textit{Layer Normalization} \cite{JimmyBa2016LayerN} before feeding the hidden states into the \textit{BiGRU}.

As described above, the forward block can be formalized as follow:
\begin{equation}
    \begin{aligned}
        H^f_l     &= W\textit{BiGRU} \left( \textit{LayerNorm} \left( H^{\prime}_l \right) \right) + b\\
        H^{\prime}_{l+1} &= \sigma \left( \textit{Conv1D} \left( H_l \right) \right)\\
    \end{aligned}
\end{equation}
where $\sigma$ is activate function which we exploit the $\textit{Gelu}$ \cite{hendrycks2016gaussian} in our work.
In particular, there is $H^{\prime}_0=\left[h_0,h_1,h_2,...,h_L\right]$, namely the initial inputs to the pyramid network is exactly the output from the sentence encoder.

\subsubsection{Backward Block}
The backward block allows layers to receive feedback from the higher-level neighbors,  which benefits the model to learn better representations at each layer by enhancing the communication between the neighbor tokens.
In bottom-to-up propagation, the sequence reduces the length each time passes through the \textit{CNN}.
Thus the corresponding \textit{Transposed CNN}s are required to reconstruct the representation of the text span embeddings.

Specifically, in the $l-1$-th inverse layer, we first feed the hidden states transmitted down from the $l$-th  layer into \textit{BiGRU} to catch the spans' interaction.
And then, we concatenate the outputs with the outcomes from the $l-1$-th forward layer.
The concatenated representations are linear transformed and then summed up with the pooled residuals to get the outcomes at the $l-1$-th layer.
The outcomes will be passed through the $\textit{Transposed CNN}$ and then sent to the next layer.
Similarly, \textit{Layer Normalization} is deployed due to the diversity of input scales.

The overall structure can be formalized as follows:
\begin{equation}
    \begin{aligned}
        H^{\prime}_{l-1} &= H^{f}_{l-1} \oplus \textit{BiGRU}\left(\textit{LayerNorm}\left(H^b_{l}\right)\right)\\
        H_{l-1} &= \textit{LayerNorm}\left(\textit{Pool}\left(H\right)+W H^{\prime}_{l-1} + b\right)\\
        H^b_{l-1} &= \sigma\left(\textit{TransposedConv1D}\left(H_{l-1}\right)\right)\\
    \end{aligned}
\end{equation}
where $\sigma$ is activate function and $\textit{Gelu}$ is utilized. 
$\oplus$ denotes the concatenation operation.
Here, the length of the residuals and the feature representations of the current layer were aligned using average pooling.
The purpose of designing this residual connection is to avoid model degradation with the increase of the feature pyramid layers.

\subsubsection{Proposal Fuse}
After the build-up of the feature pyramid, we are able to initialize the entity proposals.
In contrast to the works \cite{DianboSui2020JointEA, ijcai2021-542} which manually create the query vectors, the encoder-only architecture is deployed in our work in which query vectors also work as semantic feature vectors. 

Note that the entity proposal mentioned in this paper includes not only query vector $q_i \in \mathbb{R}^d$ but also category logarithms $c_i \in \mathbb{R}^{T+1}$ and span location $n_i \in \mathbb{R}^2$, where $d$ stands for the dimension of queries vector and $T$ represents the number of the pre-defined categories.
Additional one for $T+1$ corresponds to the $\textbf{None}$ type.
The span location indicates the start and end position of the entity.
The proposals are initialized with the help of the prepared feature pyramid.

Apparently, each feature vector $h_{i,l}$ in the pyramid specifically corresponds to a text span $n_{i,l}=\left[s_{i,l},e_{i,l}\right]$.
Our model computes the scores vector $r_{i,l}$ for each feature and utilizes them to calculate the weights vector for each proposal as follows:  
\begin{equation}
    \begin{aligned}
        r_{i,l} &= \textit{MLP}\left(h_{i,l}\right) \\
            &= \left[r^s_{i,l},r^e_{i,l}\right]\\
        \alpha^j_{i,l} &= \frac{\textit{exp}\left(r_{i,l}\right)}{\sum\limits_{\left(i^{\prime},l^{\prime}\right)\in \mathcal{N}_j}\textit{exp}\left(r_{i^{\prime},l^{\prime}}\right)}
    \end{aligned}
    \label{proposer_weight}
\end{equation}
where $\alpha^j_{i,l} \in {[0,1]}^2$ denotes the weight vector of the span at index $i$ of the $l$-th level for $j$-th proposal that corresponds to the token $w_j$.
$\mathcal{N}_j = \left\{ \left(i,l\right)\vert j\in \left[s_{i,l},e_{i,l}\right] \right\}$ present positions of features whose corresponding text span contains the token $w_j$.
The coefficients of aggregation at the start and end positions are calculated separately, which can enhance the expressiveness of the aggregation process and increase the range of the initial span.

With the weight vector, the location span $n_j$ can be initialized as follows:
\begin{equation}
    n_j = \sum\limits_{\left(i,l\right)\in \mathcal{N}_j}\alpha^j_{i,l} \cdot p_{i,l}
    \label{proposer_aggregation}
\end{equation}
where $\cdot$ stands for the element-wise product and $p_{i,l}$ represent the span location correspond to the feature $h_{i,l}$.
One may notice that the position of the initial span is actually bounded by the leftmost span and the rightmost span containing the token $j$.
Such limitation helps the model to give more reasonable proposals.

As for query vector $q_i$, we directly employ the features at the bottom layer, namely $h_{i,1}$.
The reason for doing so is that the features at the bottom layer are obtained from the whole propagation, which means they contain the most semantic messages of the pyramid.
Once obtain the query vector, the category logarithms $c_j$ can be naturally derived using a multi-layer perceptron:
\begin{equation}
    c_j = \textit{MLP}\left(q_j\right) \\
\end{equation}
The category logarithms will be further utilized and corrected in the regressor network.

\subsection{Regressor}
\begin{figure}
    \centering
    \includegraphics[width=0.75\textwidth]{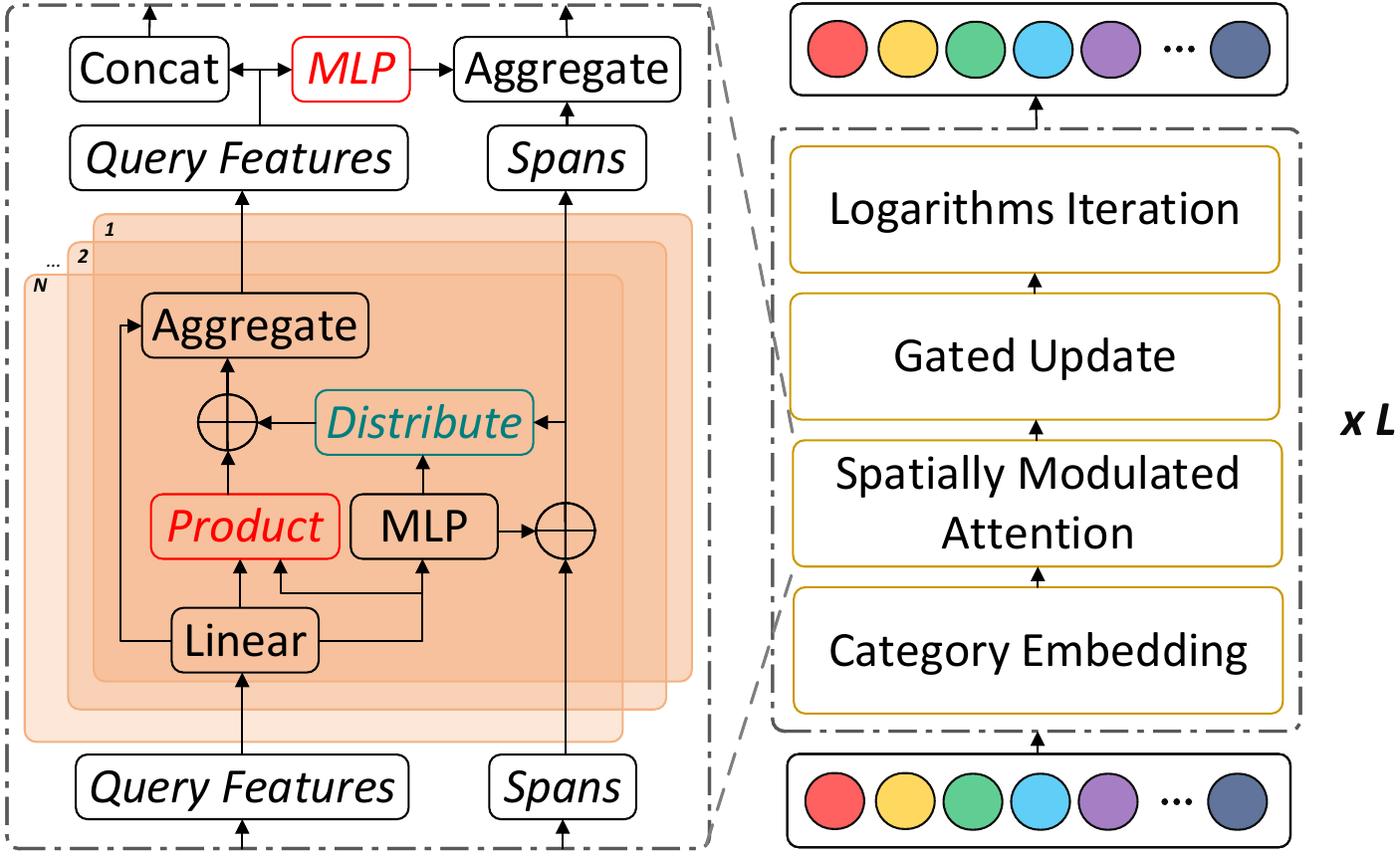}
    \caption{General view of regressor and detailed structure of spatially modulated attention.}     
    \label{regressor}
\end{figure}

Previous works \cite{ZhiqingSun2020RethinkingTS, ZihangJiang2020ConvBERTIB} have proven the importance of the sparsity of the attention map in accelerating the training.
Based on their conclusion, we introduce the spatially modulated attention in the regressor.
We employ the encoder-only architecture in which semantic feature vectors directly play the role of the query instead of manually creating them. 
Inspired by successful practice on auxiliary loss \cite{10.1007/978-3-030-58452-8_13, RamiAlRfou2018CharacterLevelLM}, we further introduce progressive refinement into the regressor. 
The overall architecture of the regressor is shown in Figure \ref{regressor}.

\subsubsection{Category Embedding}
Each proposal is inputted alongside category logarithms, which carry the semantic information of the proposal category.
We follow the way of embedding the positional message in the \textit{Transformer} \cite{2017Attention} using the element-wise sum to integrate the query vector with the category information:
\begin{equation}
    \begin{aligned}
        \alpha_i &= \textit{Softmax}\left(c_i\right)\\
        H^c &= \left[h^c_1,h^c_2,...,h^c_{T+1}\right]\\
        h^{\prime}_i &= q_i + \alpha^T_i H^c
    \end{aligned}
    \label{category_embed}
\end{equation}
where $\alpha_i$ stands for the category weights vector of $i$-th proposal.
$H^c \in \mathbb{R}^{d \times \left(T+1\right)}$ represents the type embedding matrix.
$h^{\prime}_i$ denotes the category embedded query vectors.

In an encoder-only design, query vectors are also semantic feature vectors, making it difficult for the model to discriminate the role played by each query vector.
We created category embeddings based on this understanding to clarify the function of each query vector in the forward propagation process.

\subsubsection{Spatially Modulated Attention}
The central idea of the spatially modulated attention mechanism is to utilize the spatial guidance from the span locations of each entity proposal to learn the reasonable sparse attention maps.

\paragraph{\textbf{Dynamic spatial weight maps}}
A Gaussian-like distribution is generated for each proposal in each attention head. 
The Gaussian-like distribution in this paper is given as:
\begin{equation}
    \begin{aligned}
        G\left(x\right) = e^{\left(x-\mu\right)^T\Theta\left(x-\mu\right)}
    \end{aligned}
\end{equation}
where $x \in \mathbb{R}^2$ denotes independent variable.
$\mu \in \mathbb{R}^2$ can be seen as the expectation vector, which represents the center of the distribution.
$\Theta \in \mathbb{R}^{2 \times 2}$ can be seen as the precision matrix (namely the inverse of the covariance matrix $\Sigma^{-1}$), which described the shape of the distribution.

The main difference between the Gaussian-like distribution and the multidimensional Gaussian distribution is that we remove the probability normalization term.
The main reason for doing so is twofold.
First, the normalized coefficients of the multidimensional Gaussian distribution are derived in the continuous case.
Whereas the case here is a discretized attention map.
Second, the subsequent \textit{Softmax} operation will also do the normalization work.

Sets of parameters are needed to generate the distributions.
We formulate our ways of producing them as follows:
\begin{equation}
    \begin{aligned} 
        \Delta n_{i,m},\theta_{i,m} &= \textit{MLP}\left(h^q_{i,m}\right)\\
        \mu_{i,m} &= n_i + \Delta n_{i,m}\\
        \Theta_{i,m} &= \theta_{i,m}\theta_{i,m}^T \\
    \end{aligned}
    \label{distribute}
\end{equation}
where $h^q_{i,m}$ is the linear transformation for $m$-th head of $h^{\prime}_i$ in Formula \ref{category_embed}, which is also the query in the \textit{QKV Attention} \cite{2017Attention}.
$\Delta n_{i,m}$ denotes the difference between the span location of the proposal and the center of the map.
From Formula \ref{distribute}, it can be found that $\Theta_{i,m}$ is a symmetric positive definite matrix satisfying the requirement of being the precision matrix for a Gaussian distribution.

It can be observed that the model generates the spatial weight map dynamically, implying that we expect the model to learn the way to make better use of the spatial prior.

\paragraph{\textbf{Spatially modulated attention}}
The spatially modulated attention mechanism aims to ease the difficulty of learning the proper attention pattern by enhancing the sparsity of the attention map with the spatial weight map.
With the dynamically generated spatial prior $G$, the spatially modulated attention can be conducted as follows:
\begin{equation}
    \begin{aligned}
        r^m_{i,j} &= \frac{\left(W^q_m h^{\prime}_i + b^q_m\right)^T \left(W^k_m h^{\prime}_i + b^k_m\right)}{\sqrt{d/M}} + \log{G_{i,m}\left(n_j\right)}\\
        \alpha^m_{i,j} &= \frac{\textit{exp}\left(r^m_{i,j}\right)}{\sum\limits_{j^{\prime}=1}^L \textit{exp}\left(r^m_{i,j^{\prime}}\right)}\\
        o_{i,m} &= \sum\limits_{j=1}^L \alpha^m_{i,j}\left(W^v_m h^{\prime}_j + b^v_m\right)
    \end{aligned}
\end{equation}
where $m$ denotes the $m$-th attention head, $M$ stands for the number of heads.
The spatially modulated attention mechanism performs the element-wise sum between the spatial graph and the dot product attention and then undertakes the $\textit{Softmax}$ normalization. 
By introducing the spatial graph, each query vector has a relatively high weight for the query vector whose corresponding span location is closer.
It limits the search space for attention patterns and thus reduces the difficulty of the convergence.

\paragraph{\textbf{Multi-head iteration}}
We aggregate the results of each attention head to produce new query vectors and span locations.
This process is similar to the way the proposer fuses the proposals.
First, we compute the weight vectors for each head:
\begin{equation}
    \begin{aligned}
        r_{i,m} &= \textit{MLP}\left(o_{i,m}\right) \\
            &= \left[r^s_{i,m},r^e_{i,m}\right]\\
        \alpha_{i,m} &= \frac{\textit{exp}\left(r_{i,m}\right)}{\sum^M \limits_{m^{\prime}=1}\textit{exp}\left(r_{i,m^{\prime}}\right)}\\
        &= \left[\alpha^s_{i,m^{\prime}},\alpha^e_{i,m^{\prime}}\right]\\
    \end{aligned}
\end{equation}
where $M$ is the number of attention heads, $r_{i,m}$ denotes the score of $m$-th head in $i$-th proposal.
We calculate the corresponding score based on the aggregation result of each head and do the \textit{Softmax} normalization to get the weight vectors.

Using the weight vectors, we conduct the iteration of the span location.
Similarly, we perform aggregation for the start and end positions separately.
The procedure can be formulated as follows:
\begin{equation}
    n_i = \sum^M \limits_{m=1}\alpha_{i,m} \cdot \mu_{i,m} \\
\end{equation}
The spatially modulated attention implements iteration of span location based on the weighted average of each head's center.
Fusing iterations on span location and multi-head attention mechanisms increases the flexibility of span updates and benefits the accuracy of model predictions.

As for query vectors, we simply follow the custom way of aggregating the output of each head as follows:
\begin{equation}
    q^{\prime}_i = \sum^M \limits_{m=1} \left(W_m o_{i,m} + b_m\right)\\
    \label{attention_output}
\end{equation}

\subsubsection{Gated Update}
We replace the feed-forward module in \textit{Transformer} with the gate mechanism in the GRU \cite{cho-etal-2014-learning} to strengthen the expressive ability of the model.
The gate mechanism can be formulated as follows:
\begin{equation}
    \begin{aligned}
        r &= \textit{Sigmoid}(W_{xr}x+b_{xr}+W_{hr}h+b_{hr})\\
        z &= \textit{Sigmoid}(W_{xz}x+b_{xz}+W_{hz}c+b_{hz})\\
        n &= \textit{Tanh}(W_{xn}x+b_{xn}+r\odot(W_{hn}h+b_{hn}))\\
        o &= (1-z)\odot n+z\odot h\\\\
    \end{aligned}
\end{equation}
where $x$ is the current input, $h$ is the hidden state, and $o$ is the final new state output.

We denote the above mechanism as $\text{Gate}(\cdot)$.
The new query vectors are obtained as follows:
\begin{equation}
q_i= \textit{Gate}(h_i^{\prime}, q_i^{\prime})
\end{equation}
Where $q_i^{\prime}$ stands for the outputs from the spatially modulated attention in Formula \ref{attention_output} and $h_i^{\prime}$ is category embedded represent of the query in Formula \ref{category_embed}.
The attention mechanism's outcome is used as the input to update the query representations in each iteration, with the original state representations serving as the hidden states.

\subsubsection{Logarithms Iteration}
After updating the query vector $q_i$ and span locations $n_i$, we further iterative the category logarithms $c_i$.
Same as in span location, the module predicts the difference of logarithms rather than directly outputting the result.
The updating procedure can be formulated as follows:
\begin{equation}
    \begin{aligned}
        \Delta c_i &= \textit{MLP}\left(q_i\right) \\ 
        c_i &= c_i + \Delta c_i
    \end{aligned}
\end{equation}
The assumption that each entity proposal may change responsibilities during the iteration is the reason for undertaking logarithms iterations.

\subsection{Prediction Head}
The prediction head produces predictions based on the joint probability distribution indicated by entity proposals.
First, we can use \textit{Softmax} normalization to convert the category logarithms into the probability distribution:
\begin{equation}
    \begin{aligned}
        \Delta c_i &= \textit{MLP}\left(q_i\right) \\ 
        p^c_i &= \textit{Softmax}\left(c_i + \Delta c_i\right)
    \end{aligned}
\end{equation}
As for span location distribution, it is natural to extend the mechanism of spatially modulated attention as the joint pointer network, which is similar to the fusion of the smooth boundary \cite{EnweiZhu2022BoundarySF} and the biaffine decoder \cite{yu-etal-2020-named}.
Specifically, we calculate the joint probability distribution of the span location in the following way:
\begin{equation}
    \begin{aligned}
        g_i\left(s,e\right) &= \left(W_s q_i + b_s \right)^T \left(W_s h_s + b_s \right) + \left(W_e q_i + b_e \right)^T \left(W_e h_e + b_e \right)\\
        r_i\left(s,e\right) &= \log{G_i\left(s,e\right)} + g_i\left(s,e\right)/\sqrt{d} \\
        p^n_i &= \textit{Softmax}\left(\{r_i(s,e)\vert \left( s \le e \right) \wedge \left(s,e \le L \right)\}\right)
    \end{aligned}
\end{equation}
where $G$ is created in the same way as spatially modulated attention. 
$g_i$ is the score derived from the joint pointer network, which is essentially the sum of two pointer networks.

\subsubsection{Training}
In the training phase, the determination of the best match for each proposal is a one-to-many Linear Assignment Problem.
We formulate the search procedure as follows:
\begin{equation}
    \hat{\beta} = \mathop{\textit{argmin}} \limits_{\beta \in \mathcal{O}_L} \sum \limits_{i=1}^L \mathcal{L}_{\textit{match}}\left(\hat{\mathcal{Y}}_{\beta\left(i\right)}, \mathcal{Y}^{\prime}_i\right)
    \label{assign}
\end{equation}
where $\hat{\mathcal{Y}} = \mathcal{Y} \cup \{\left(\varnothing,\varnothing,\varnothing \right)\}$ denotes the targets with padding triple denoting \textbf{None}.
$\mathcal{Y} = \{\left(s_i,e_i,t_i\right)\}_{i=1}^N$ denotes the set of entities existing in the sentence.
$\mathcal{Y}^{\prime}$ signifies the proposals outputted by the model, in which each element can be represented as $\left(p^c,p^n\right)$.
The padding in $\hat{\mathcal{Y}}$ is necessary as there is the possibility is no entity in the sentence, in which case all the proposals should target at None.
The $\mathcal{L}_{\textit{match}}$ is calculated in the following way:
\begin{equation}
     \mathcal{L}_{\textit{match}}\left(\hat{\mathcal{Y}}_{\beta\left(i\right)}, \mathcal{Y}^{\prime}_i\right)= -\log{p^c_i\left(t_{\beta\left(i\right)}\right)} - \mathbb{I}_{t_{\beta\left(i\right)} \ne \varnothing}\left[ \log{p^n_i\left(s_{\beta\left(i\right)},e_{\beta\left(i\right)}\right)} \right]
\end{equation}

$\mathcal{O}_L$ in Formula \ref{assign} stands for the assignment mappings satisfying certain constraints when the number of proposals is $L$.
For the circumstance when $L > N$, any $\beta \in \mathcal{O}_L$ has to be surjective mapping, which means every candidate entity will be assigned to a proposal.
As for $L \le N$, the mapping $\beta$ ensure $\hat{\mathcal{Y}}_{\beta\left(i\right)} \in \mathcal{Y}$, namely any proposal will be assign to an entity.
Given the constraints and optimization objectives, we use Algorithm \ref{assignment} to solve the assignment problem.

\begin{algorithm}
	\caption{Assignment}
	\label{assignment}
	\begin{algorithmic}[1]
	    \Require{Targets $\hat{\mathcal{Y}}$, Proposals $\mathcal{Y}^{\prime}$}
		\For{$i = 1$ \textbf{to} $L$}
		    \For{$j = 1$ \textbf{to} $N$}
		        \State $C_{i,j} \gets \mathcal{L}_{\textit{match}}\left(\hat{\mathcal{Y}}_j,\mathcal{Y}^{\prime}_i\right)$
		    \EndFor
		\EndFor
		\State $\hat{\beta} \gets \textit{HUG}(C)$
	    \For{$i$ \textbf{in} $\left\{j\vert \mathcal{Y}^{\prime}_j \textit{ haven't been assigned in }\hat{\beta} \right\}$}
	        \State $k \gets \textit{argmin}\left(C_i\right) $
	        \State $\hat{\beta} \gets \hat{\beta} \cup \{\mathcal{Y}^{\prime}_i \xrightarrow{} \hat{\mathcal{Y}}_k \} $
	    \EndFor
	\end{algorithmic}  
\end{algorithm}

In the algorithm, $C$ stands for the cost matrix recording the matching cost of each proposal-entity pair. 
So the $C_{i,j}$ denotes the matching cost between $\mathcal{Y}^{\prime}_i$ and $\hat{\mathcal{Y}}_j$ and $C_i$ represents a vector in which element is the cost between $\mathcal{Y}^{\prime}_i$ and a target in $\hat{\mathcal{Y}}$.
\textit{HUG} denotes the Hungarian algorithm \cite{HaroldWKuhn1955TheHM}, which can solve the assignment problem given $L \le N$.
We use the greedy strategy when $L>N$, which means the entity with the minimal matching loss (including padding) for each unassigned proposal in the Hungarian algorithm will be the corresponding prediction target.

With the optimal pairing $\hat{\beta}$, the final bipartite loss function is simply the matching cost:
\begin{equation}
    \begin{aligned}
        \mathcal{L}\left(\hat{\mathcal{Y}}, \mathcal{Y}^{\prime}\right) = \sum \limits_{i=1}^L \mathcal{L}_{\textit{match}}\left(\hat{\mathcal{Y}}_{\hat{\beta}\left(i\right)}, \mathcal{Y}^{\prime}_i\right)
    \end{aligned}
\end{equation}

\subsubsection{Predict}
In the prediction phase, we utilize the probabilities distribution outputted by the model to produce the final results through  Algorithm \ref{predict}.

\begin{algorithm}
	\caption{Prediction}
	\label{predict}
	\begin{algorithmic}[1]
	    \Require{Proposals $\mathcal{Y}^{\prime}$}
	    \State $\mathcal{E} \gets \left\{\right\}$
		\For{$\left(p^c,p^n\right)$ \textbf{in} $\mathcal{Y}^{\prime}$}
		    \State $t \gets \textit{argmax}(p^c)$
		    \State $(s,e) \gets \textit{argmax}(p^n)$
		    \If{$t \ne \varnothing \wedge p^c\left(t\right) \cdot p^t\left(s,e\right) \ge  p^c\left(\varnothing\right)$ }
		        \State $\mathcal{E} \gets \mathcal{E} \cup \left(s,e,t\right)$
		    \EndIf
		\EndFor
	\end{algorithmic}  
\end{algorithm}

We examine each proposal to see if it is more likely to represent the entity than to be the padding.
In particular, we assume that the probability that a proposal represents an entity is the product of its corresponding category probability and the localization probability, and the probability that it is the padding is the probability that its category is \textbf{None}.
So a proposal is considered to signify an entity when and only when the maximum of its span location probability times the maximum of its category probability overpasses its probability of being \textbf{None}.

\section{Experiments}
The datasets, baselines, and settings used in our experiments are described in this section.
We perform a series of experiments to illustrate the characters of our model.
Table \ref{statistic} displays the statistics of the datasets.

\begin{table}
\centering
\caption{Statistical information on English datasets. \#S: the number of sentences. \#AS: the average length of sentences. \#NE: the number of entities. \#AE: the average length of entities.  \#ME: the maximum number of entities in a sentence.}
\label{statistic}
\begin{tabular}{lccclccclccc}
\toprule
\multirow{2}{*}{Info.} & \multicolumn{3}{c}{GENIA} &  & \multicolumn{3}{c}{CoNLL03} &  & \multicolumn{3}{c}{WeiboNER} \\ \cline{2-4} \cline{6-8} \cline{10-12} 
                       & Training   & Test   & Dev    &  & Training    & Test    & Dev    &  & Training    & Test    & Dev     \\ \midrule
\#S                    & 15203   & 1669   & 1854   &  & 14041    & 3250    & 3453   &  & 1350     & 270     & 270     \\
\#AS                   & 25.4    & 24.6   & 26.0   &  & 14.5     & 15.8    & 13.4   &  & 33.6     & 33.2    & 33.8    \\
\#NE                   & 46142   & 4367   & 5506   &  & 23499    & 5942    & 5648   &  & 1834     & 371     & 405     \\
\#AE                   & 1.94    & 2.14   & 2.08   &  & 1.43     & 1.43    & 1.42   &  & 1.24     & 1.19    & 1.20    \\
\#ME                   & 21      & 14     & 25     &  & 20       & 31      & 20     &  & 17       & 13      & 12      \\ \bottomrule
\end{tabular}
\end{table}

\subsection{Datasets}
Experiments were carried out using public datasets including GENIA \cite{10.5555/1289189.1289260}, CoNLL03 \cite{tjong-kim-sang-de-meulder-2003-introduction}, and WeiboNER \cite{peng-dredze-2015-named}.

GENIA is a biological dataset including five entity types: DNA, RNA, protein, cell lineage, and cell type.
It contains a high portion of nested entities.
The same experimental setup as \cite{yu-etal-2020-named} was used.
We did not conduct experiments on additional authoritative nested datasets in English because we lacked access to the datasets' copyright.

CoNLL03 dataset is a huge dataset frequently utilized.
The main content in the dataset is the news report from the RCV1 corpus of Reuters.
Locations, organizations, people, and information are among the named features.
The setting \cite{yan-etal-2021-unified-generative} combining the training and validation sets is followed.
We demonstrate the generalizability of our approach on flat datasets through experiments on this dataset.

The WeiboNER dataset is a Chinese dataset from the Weibo social platform.
It includes named and nominal entity types of individuals, organizations, locations, and geopolitics.
The setup of our experiments on this dataset is identical to \cite{li-etal-2020-flat}.
We use this Chinese dataset to evaluate the model's cross-language ability.

\subsection{Training Details}
We use the pre-trained \textit{BERT} model developed by the open framework Transformers \cite{ThomasWolf2019HuggingFacesTS} in our experiments.
In order to acquire a better contextual representation in biological domain text, we substituted \textit{BERT} with \textit{BioBERT} \cite{10.1093/bioinformatics/btz682} for the GENIA dataset and used the \textit{BIO-word2vec} \cite{chiu-etal-2016-train} to decrease the quantity of out-of-vocabulary words.
We use \textit{Glove} \cite{pennington-etal-2014-glove} as pre-trained word vectors for the English dataset.
We use word vectors developed by \cite{li-etal-2018-analogical} for the Chinese dataset.
The models are trained by the \textit{AdamW} optimizer with the learning rate of $2e-5$.
The max gradient normalization is $1e0$ in all experiments.
The learning rate is altered along with the training procedure under the cousin warm-up-decay learning rate schedule.

\subsection{Comparison}
We compare our model to many powerful state-of-the-art models for the GENIA dataset, including \textbf{LocateAndLabel} \cite{shen-etal-2021-locate}, \textbf{Pyramid} \cite{wang-etal-2020-pyramid}, \textbf{BARTNER} \cite{yan-etal-2021-unified-generative}, \textbf{SequenceToSet} \cite{ijcai2021-542}, \textbf{NER-DP} \cite{yu-etal-2020-named}, \textbf{BioBART} \cite{https://doi.org/10.48550/arxiv.2204.03905}, \textbf{LogSumExpDecoder} \cite{wang-etal-2021-nested}.
The above baselines' reported results are taken straight from the original published literature.

\begin{table}
\caption{The performance on the GENIA dataset.}
\centering
\label{result_genia}
\begin{tabular}{lccc}
\hline
\toprule
\multirow{2}{*}{Model}                                  & \multicolumn{3}{c}{GENIA}                        \\ \cline{2-4} 
                                                        & Prec.          & Rec.           & F1             \\ \midrule
LocateAndLabel\cite{shen-etal-2021-locate}              & 80.19          & \textbf{80.89} & 80.54          \\
Pyramid\cite{wang-etal-2020-pyramid}                    & 79.45          & 78.94          & 79.19          \\
BARTNER\cite{yan-etal-2021-unified-generative}          & 78.89          & 79.60          & 79.23          \\
SequenceToSet\cite{ijcai2021-542}                       & \textbf{82.31} & 78.66          & 80.44          \\
NER-DP\cite{yu-etal-2020-named}                         & 81.80          & 79.30          & 80.50          \\
BioBART\cite{https://doi.org/10.48550/arxiv.2204.03905} & -              & -              & 79.93          \\
LogSumExpDecoder\cite{wang-etal-2021-nested}            & 79.20          & 78.67          & 78.93          \\ \hline
Our Model                                               & 81.74          & 79.76          & \textbf{80.74} \\ \bottomrule
\end{tabular}
\end{table}

Table \ref{result_genia} shows the advanced outcomes provided by our model.
Our model increases the F1 score compared with the top algorithm in the GENIA dataset.
The GENIA dataset contains numerous nested entities, where about 20\% of entities are nested.
The improvement of our model on the GENIA dataset illustrates our model's ability to extract nested entities.
Compared to the previous stare-of-art model LocateAndLabel, we avoid the enumeration of candidate spans through the employment of the set prediction network and achieve the performance.
The set prediction can be seen as the soft enumeration with the flexibility lacking in the manual preparation of the candidate spans.
Compared with another work SequenceToSet, which also achieved excellent results, we deploy the encoder-only architecture to implement the performance advancement while bypassing the difficulty of training query vectors.
The encoder-only architecture accelerates the coverage of the training.
It is shown in the following experiments that the model produces comparable results even if only trained under a small number of epochs.

For the CoNLL03 dataset, the \textbf{LocateAndLabel} \cite{shen-etal-2021-locate}, \textbf{PIQN} \cite{shen-etal-2022-piqn}, \textbf{NER-DP} \cite{yu-etal-2020-named}, \textbf{LUKE} \cite{yamada-etal-2020-luke}, \textbf{MRC} \cite{li-etal-2020-unified}, and \textbf{KNN-NER} \cite{2203.17103} are compared with our model.
The results are taken directly from the original published literature.
Note that \cite{yan-etal-2021-unified-generative} provides the results of LUKE, NER-DP, and MRC.

\begin{table}
\caption{The performance on CoNLL03 dataset.}
\centering
\label{result_CoNLL03}
\begin{tabular}{lccc}
\hline
\toprule
\multirow{2}{*}{Model}                                  & \multicolumn{3}{c}{CoNLL03}                      \\ \cline{2-4} 
                                                        & Prec.          & Rec.           & F1             \\ \midrule
LocateAndLabel\cite{shen-etal-2021-locate}              & 92.13          & \textbf{93.79} & 92.94          \\
PIQN\cite{shen-etal-2022-piqn}                          & \textbf{93.29} & 92.46          & 92.87          \\
NER-DP\cite{yu-etal-2020-named}                         & 92.85          & 92.15          & 92.50          \\
LUKE\cite{yamada-etal-2020-luke}                        & -              & -              & 92.87          \\ 
MRC\cite{li-etal-2020-unified}                          & 92.47          & 93.27          & 92.87          \\
KNN-NER\cite{2203.17103}                                & 92.82          & 92.99          & 92.93          \\ \hline
Our Model                                               & 92.86          & 93.13          & \textbf{93.00} \\ \bottomrule
\end{tabular}
\end{table}

Table \ref{result_CoNLL03} shows the outcomes.
Since CoNLL03 is a well-established dataset widely investigated, the improvement on this dataset demonstrates that our model is able to generalize well on the traditional flat dataset.
Although our model introduces a certain amount of invalid query vectors, the satisfactory performance on this dataset shows that the additional entity proposals do not affect the model's ability to recognize flat entities in general.
Compared with some recent works PIQN and KNN-NER, our model still achieves some improvement in F1.

As for the WeiboNER dataset, we compare our model to various strong baselines, including \textbf{TFM} \cite{liu_tfm_2022}, \textbf{LocateAndLabel} \cite{shen-etal-2021-locate},  \textbf{KNN-NER} \cite{2203.17103}, \textbf{SLK-NER}\cite{DouHu2020SLKNERES}, \textbf{BoundaryDet} \cite{chen-kong-2021-enhancing}, \textbf{ChineseBERT} \cite{sun-etal-2021-chinesebert} and \textbf{AESINER} \cite{YuyangNie2020ImprovingNE}.
We present outcomes published in the original literature.

\begin{table}
\caption{The performance on the WeiboNER dataset.}
\centering
\label{result_weibo}
\begin{tabular}{lccc}
\hline
\toprule
\multirow{2}{*}{Model}                                  & \multicolumn{3}{c}{WeiboNER}                     \\ \cline{2-4} 
                                                        & Prec.          & Rec.           & F1             \\ \midrule
TFM\cite{liu_tfm_2022}                                  & 71.29          & 67.07          & 71.12          \\
LocateAndLabel\cite{shen-etal-2021-locate}              & 70.11          & 68.12          & 69.16          \\
KNN-NER\cite{2203.17103}                                & \textbf{75.00} & 69.92          & 72.03          \\
SLK-NER\cite{DouHu2020SLKNERES}                         & 61.80          & 66.30          & 64.00          \\
BoundaryDet \cite{chen-kong-2021-enhancing}             & -              & -              & 70.14          \\
ChineseBERT\cite{sun-etal-2021-chinesebert}             & 68.75          & \textbf{72.97} & 71.26          \\ 
AESINER\cite{YuyangNie2020ImprovingNE}                  & -              & -              & 69.78          \\ \hline
Our Model                                               & 72.93          & 71.85          & \textbf{72.38} \\ \bottomrule
\end{tabular}
\end{table}

Table \ref{result_weibo} shows that our approach reaches a new state-of-art performance provided on the WeiboNER dataset.
It demonstrates that the advantages of our model do not limit only to the English datasets.
Even compared with some approaches, like ChineseBERT, which targets the Chinese dataset, our model still produces significant improvement.
More importantly, since the WeiboNER dataset is a few-shot dataset, the success of this dataset further demonstrates the easiness of training our model.
We claim that this excellent performance stems from the good nature of our model's emphasis on localization, which is generated by spatially modulated attention.
Unlike other datasets, the WeiboNER dataset is sampled from social platforms, where the occurrence of entities and the overall sentence meaning are relatively less closely related.
Moreover, the entity length tends to be shorter due to the nature of the Chinese language, as seen in Table \ref{statistic}.
Spatially modulated attention in the model introduces the Gaussian-like distribution to increase the sparsity of the attention graph by focusing more on local information, which caters well to the characteristics of the data set and thus achieves excellent results.

\subsection{Detailed Result}
As indicated in Table \ref{entity_length}, we also looked at the performance of entities of various lengths.
Our model produces the best results when the entity length is $1$ or $2$ on all datasets.
And the F1 score gradually decreases as the length of the entity increases.

\begin{table}
\centering
\caption{Result on entities of different lengths.}
\label{entity_length}
\resizebox{\textwidth}{!}{
\begin{tabular}{lccccccccccc}
\toprule
\multicolumn{1}{c}{\multirow{2}{*}{Lenz.}} & \multicolumn{3}{c}{GENIA}                        & \multicolumn{1}{l}{} & \multicolumn{3}{c}{CoNLL03}                      & \multicolumn{1}{l}{} & \multicolumn{3}{c}{WeiboNER}                     \\ \cline{2-4} \cline{6-8} \cline{10-12} 
\multicolumn{1}{c}{}                        & Prec.          & Rec.           & F1             &                      & Prec.          & Rec.           & F1             &                      & Prec.          & Rec.           & F1             \\ \midrule
$l=1$                                       & \textbf{86.59} & \textbf{80.46} & \textbf{83.41} &                      & 93.55          & 92.68          & 93.11          &                      & \textbf{80.52} & \textbf{74.61} & \textbf{77.46} \\
$l=2$                                       & 83.01          & 80.19          & 81.58          &                      & \textbf{94.04} & 94.32          & \textbf{94.18} &                      & 57.31          & 61.84          & 59.49          \\
$l=3$                                       & 77.57          & 78.07          & 77.82          &                      & 90.04          & 91.22          & 90.63          &                      & -              & -              & -              \\
$l=4$                                       & 75.49          & 78.40          & 76.92          &                      & 92.10          & \textbf{94.59} & 93.33          &                      & -              & -              & -              \\
$l\ge 5$                                    & 72.99          & 78.99          & 75.87          &                      & 85.18          & 92.00          & 88.46          &                      & -              & -              & -              \\ \hline
All                                         & 81.74          & 79.76          & 80.74          &                      & 92.86          & 93.13          & 93.00          &                      & 72.93          & 71.85          & 72.38          \\ \bottomrule
\end{tabular}
}
\end{table}

It is under expectation because short entities can be detected immediately by the proposer, but long entities need the regressor to perform many regressions to identify.
The sensitivity of the model's performance to the entity length is a side consequence of our suggested method's efficacy.
The effect of entity length on recognition performance is also directly influenced by the different kernel sizes of the proposer.
From Formula \ref{proposer_weight} and \ref{proposer_aggregation}, it can be seen that kernel sizes directly affect the expected length of the model output entities.
The expected length in experiments mostly does not exceed $2$, and thus the results in the table are obtained quite reasonably.

\subsection{Analysis and Discussion}
In this part, we will examine how each module and its corresponding settings affect the algorithm as well as the potential causes.
\subsubsection{Ablation Study}
On the GENIA dataset, we examined how different modules in the model contributed.
Firstly, we deleted the backward module from the feature pyramid to see how it affected communication across features of different layers.
Secondly, we tested the importance of spatially prior information by removing spatial modulation from the attention mechanism.
Thirdly, we remove the gated update to explore the importance of the gate mechanism in enhancing model expressive ability.
Moreover, we remove the category embedding in order to investigate its significance in clarifying the role of each query vector.
Finally, we investigated the impact of logarithm and location iterations on the model's efficacy.
The results are shown in Table \ref{ablation_study}.

\begin{table}
\centering
\caption{Ablation study. '-' means remove the module or substitute it with another module.}
\label{ablation_study}
\resizebox{\textwidth}{!}{
\begin{tabular}{lccclccclccc}
\toprule
\multicolumn{1}{c}{\multirow{2}{*}{Model}} & \multicolumn{3}{c}{GENIA} &  & \multicolumn{3}{c}{CoNLL03} & \multicolumn{1}{c}{} & \multicolumn{3}{c}{WeiboNER} \\ \cline{2-4} \cline{6-8} \cline{10-12} 
\multicolumn{1}{c}{}                       & Prec.   & Rec.   & F1     &  & Prec.   & Rec.    & F1      &                      & Prec.    & Rec.    & F1      \\ \midrule
Origin                                     & 81.50   & 79.07  & 80.27  &  & 92.48   & 93.00   & 92.74   &                      & 72.93    & 71.85   & 72.38   \\
-Backward block                            & 80.58   & 79.65  & 80.11  &  & 91.72   & 92.66   & 92.19   &                      & 71.42    & 70.37   & 70.89   \\
-Spatial modulation                        & 81.46  & 78.71  & 80.06  &  & 92.00   & 92.95   & 92.47   &                      & 71.50    & 69.38   & 70.42   \\
-Gated Update                              & 80.85   & 78.78  & 79.80  &  & 92.36   & 93.11   & 92.73   &                      & 71.18    & 71.35   & 71.27   \\
-Category embedding                        & 81.39   & 78.58  & 79.96  &  & 92.20   & 92.79   & 92.49   &                      & 69.64    & 67.40   & 68.50   \\
-Locations iteration                       & 81.62   & 78.67  & 80.12  &  & 92.02   & 93.18   & 92.60   &                      & 71.92    & 70.86   & 71.39   \\
-Logarithms iteration                      & 81.07   & 79.11  & 80.08  &  & 92.05   & 92.66   & 92.35   &                      & 73.76    & 70.12   & 71.89   \\ \bottomrule
\end{tabular}
}
\end{table}

The reduction in performance after removing the backward block demonstrates its use in integrating the features of different scales.
In addition, since we use the bottom-level features of the feature pyramid as the query vectors used in the subsequent modules, removing the backward module reduces the depth of the network to some extent and weakens the ability of the model to express semantics.

The decrease in F1 scores after taking out spatial modulation emphasizes the need for guidance of spatial prior knowledge.
Spatial modulation contributes to the sparsity of the attention map and facilitates the learning of proper attention patterns.
It can be observed that the performance drop after removing spatial modulation on the WeiboNER dataset is far more significant than that coming on the other datasets, which is closely related to the nature of the dataset that focus more on local messages.

The reduction of the F1 score after removing the gate mechanism proves it helps improve the model's expressive ability.
It actually plays the same role as the feed-forward layer in the \textit{Transformer}, i.e., it increases the capacity and nonlinearity of the model.

The performance deterioration after removing category embedding demonstrates the importance of explaining the role of each query vector.
Because of the encoder-only architecture used in the proposed approach, the query vector also functions as a semantic vector.
This architecture has the potential to introduce confusion about the roles of the query vectors, which is well mitigated by category embedding.

The F1 value reduces when logarithm and location iterations are removed.
It emphasizes the significance of iteration in the regression network.
We claim it plays a similar role to auxiliary loss \cite{10.1007/978-3-030-58452-8_13} and box refinement \cite{ZacharyTeed2022RAFTRA}.
In the auxiliary loss, the output results of each layer are decoded and the loss is calculated, thus enabling supervised learning for each layer.
In the box refinement, the detection box is gradually adjusted until the eventual prediction is made, which is similar to the iterative process in this paper.
Both of them apply the direct updates gradient to each layer, while the latter does the prediction based on the combination of outputs from all layers.
These strategies can reduce the difficulty of model training on the one hand.
On the other hand, it improves the performance of the model.

\subsubsection{Impact of Kernel Sizes} 
We explore the effect of kernel sizes in the pyramid and report the results in Table \ref{kernel_sizes}. 
We tested different layers of the pyramid and various combinations of kernel sizes.
All models are trained with $5$ epochs, with the number of heads and layers equal to $8$ and $3$.

\begin{table}
\centering
\caption{Impact of kernel sizes. $[2,3]$ represents a three-layer feature pyramid with two convolution kernels of size $2$ and $3$. Other symbols indicate structures in the same way. In particular, $[]$ indicates that there is no convolution kernel in the proposer module.}
\label{kernel_sizes} 
\resizebox{\linewidth}{!}{
\begin{tabular}{lccccccccccc}
\toprule
\multicolumn{1}{c}{\multirow{2}{*}{Kernel Sizes}} & \multicolumn{3}{c}{GENIA}                        &  & \multicolumn{3}{c}{CoNLL03}                      &  & \multicolumn{3}{c}{WeiboNER}                     \\ \cline{2-4} \cline{6-8} \cline{10-12} 
\multicolumn{1}{c}{}                              & Prec.          & Rec.           & F1             &  & Prec.          & Rec.           & F1             &  & Prec.          & Rec.           & F1             \\ \midrule
{[}{]}                                            & 81.35          & 78.60          & 79.95          &  & 92.21          & 92.95          & 92.58          &  & 73.35          & 71.35          & 72.34          \\
{[}2{]}                                           & 80.88          & 78.44          & 79.69          &  & 92.34          & 92.88          & 92.61          &  & 72.44          & 70.12          & 71.26          \\
{[}2,2{]}                                         & \textbf{81.50} & 79.07          & \textbf{80.27} &  & \textbf{92.48} & \textbf{93.00} & \textbf{92.74} &  & 72.58          & 68.64          & 70.55          \\
{[}2,2,2{]}                                       & 81.23          & 78.55          & 79.87          &  & 91.49          & 92.77          & 92.13          &  & \textbf{72.93} & \textbf{71.85} & \textbf{72.38} \\
{[}2,2,2,2{]}                                     & 81.24          & 78.67          & 79.94          &  & 92.24          & 92.63          & 92.43          &  & 72.19          & 69.87          & 71.01          \\
{[}2,3{]}                                         & 80.09          & \textbf{79.09} & 80.01          &  & 92.00          & 92.74          & 92.37          &  & 72.68          & 71.60          & 72.13          \\
{[}2,3,2{]}                                       & 81.19          & 78.35          & 79.74          &  & 91.75          & 92.45          & 92.10          &  & 71.89          & 71.35          & 71.62          \\ \bottomrule
\end{tabular}
}
\end{table}

The model reaches satisfactory performance when kernel sizes equal $[2,2]$ or $[2,2,2]$.
In our opinion, this is due to the fact that under these settings' entities of various lengths are mostly covered, which allows query vectors to integrate the semantic information of tokens surrounding the proposal for more precise localization.
Additionally, the length of the proposals under these kernel settings is close to the average length of entities in the dataset, which diminishes the difficulties of model training.

In addition, we observe that the model outperforms its kernel size of $[2,2,2,2]$ even when the kernel size is $[]$.
We believe that the cause of this phenomenon is twofold.
On the one hand, different kernel sizes lead to different expected lengths of proposals, and the model will achieve better performance when the expected length of proposals is close to the average entity length.
When the kernel size is $[]$, there is no convolution kernel in the proposer module, which means all the proposals will be length $1$.
So, the expected length of the proposals will be $1$, which is closer to the average length of entities in the dataset than when the kernel size is $[2,2,2,2]$.
On the other hand, with a kernel size of $[2,2,2,2]$, the feature pyramid has more layers, and the network is deep than when the kernel size is $[]$, which can lead to a relatively more difficult training of the model.

\subsubsection{Influence of Regressor Layers} 
\begin{figure}[H]
    \centering   
	\includegraphics[width=0.75\textwidth]{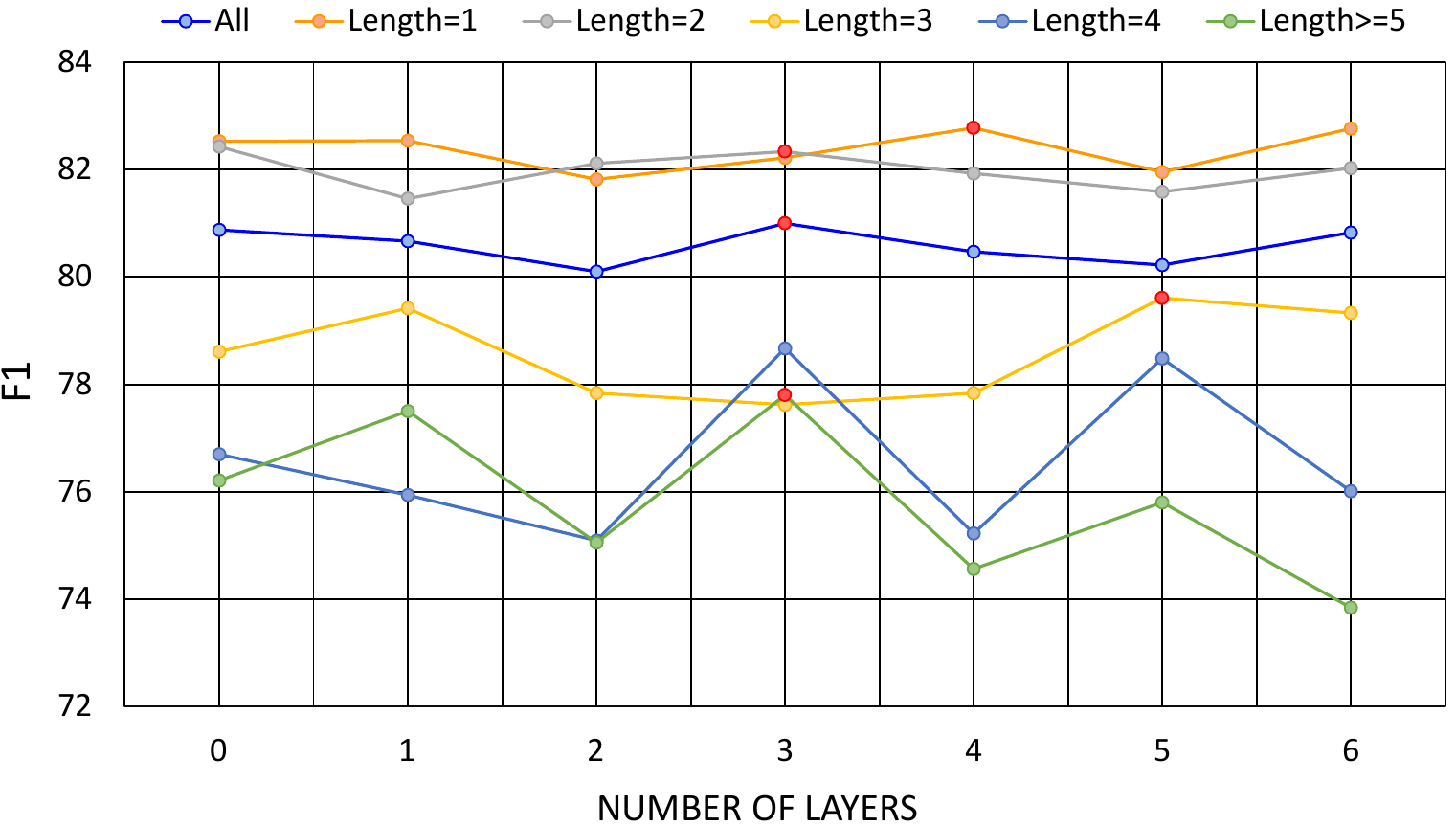}
    \caption{Result on models of different numbers of layers. The highest value of each curve is marked with a red dot.}
    \label{num_layers}
\end{figure}

To investigate the influence of the layer, we experimented with models of different layers on the GENIA dataset.
All the models were trained with $5$ epochs.
The sizes of kernels and number of heads equal $[2,3]$ and $8$ relatively.
Figure \ref{num_layers} shows the results on different entities length for the models of the different number of layers.
We can see our model reaches a comparable result when $l=3$.
To balance the time cost and model performance, we set $l=3$ for other experiments.

It can be directly observed that the recognition efficiency of short entities is less affected by the number of regression layers.
On the other hand, the recognition efficiency of entities is more sensitive to the number of regression layers.
It is because the proposer can make almost accurate proposals directly for shorter entities, while for long entities, the refinement by the regressor is necessary to identify them correctly.

\begin{figure}[H]
    \centering   	\includegraphics[width=0.75\textwidth]{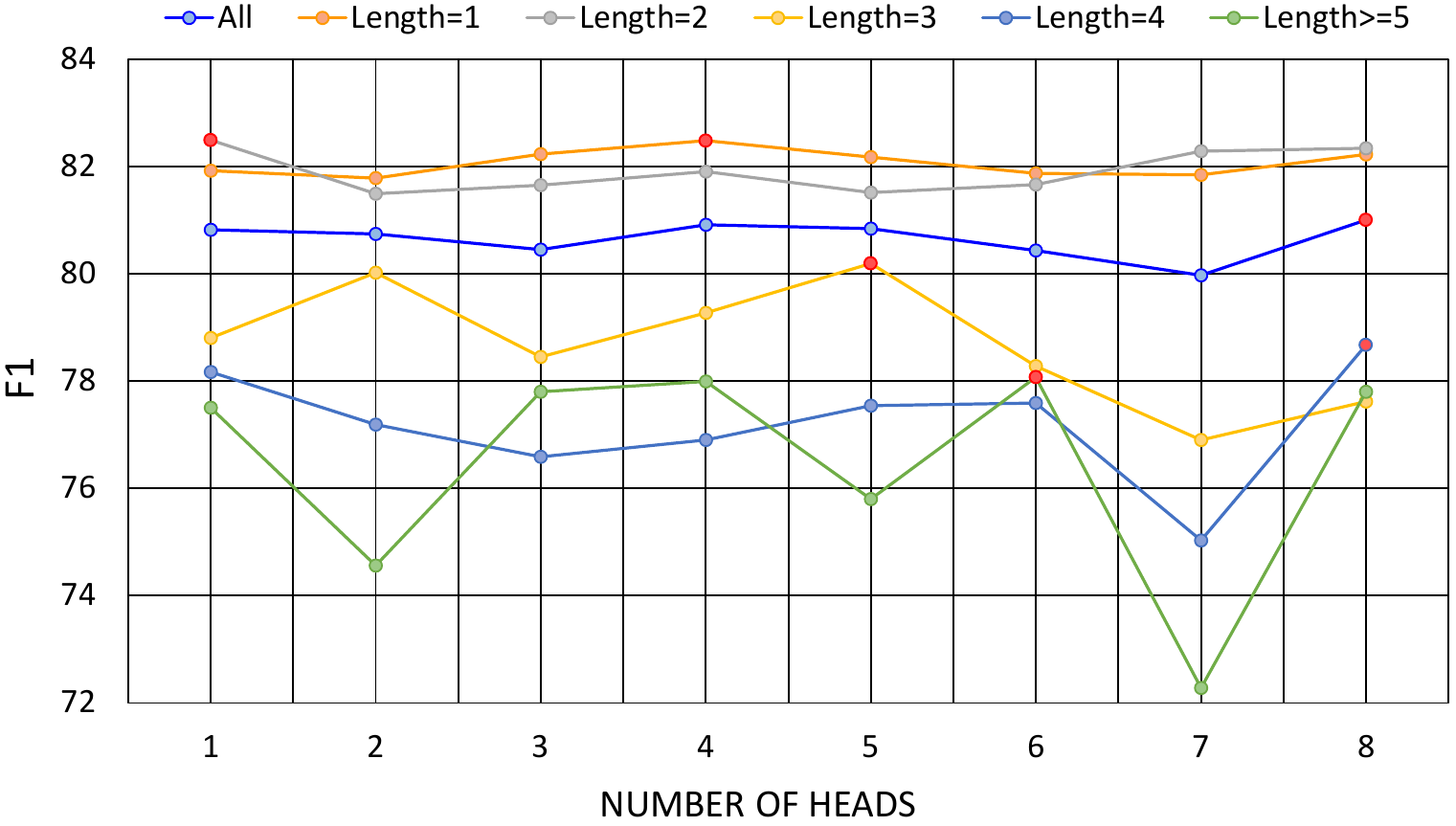}
    \caption{Result on models of different numbers of heads. The highest value of each curve is marked with a red dot.}
    \label{num_heads}
\end{figure}

\subsubsection{Analysis of Attention Heads} 
We also evaluate the model on the GENIA dataset to see how attention heads affect the results.
All of the models were trained using $5$ epochs with sizes of kernels and the number of layers equal $[2,3]$ and $3$ relatively.
The association between the number of heads and model performance is demonstrated in Figure \ref{num_heads}.
Our model produces the best results when the number of heads equals $8$.
So we set it equal to $8$ for other experiments.

Similarly, the longer the length of the entity, the more the performance of recognition is affected by the number of attention heads.
It is because the number of attention heads directly affects the iterative process of the regressor, which is indispensable for long entities.

\subsubsection{Computational Complexity} 
Theoretically, the number of intermedium proposals generated in the proposer is $O(kL)$, where $L$ stands for the length of the sentence and $k$ stands for the number of layers in the proposer.
The amount of outputted high-quality proposals is always identical to the sentence length $L$.
Compared to the traditional span-based methods classifying all the spans, which is of complexity $O(L^2)$, our proposer is relatively efficient.
While for the regressor, our model requires the computation of attention between proposals, which causes the complex of $O(nL^2)$.
The $n$ denotes the number of layers in the regressor.
On the other hand, if we employ the spatial weights as attention weights instead of fusing them with dot product weights, the theoretical complex will go down to the $O(nL)$.
Empirically, the computation of the proposer and regressor with a proper implementation is highly parallel and much lighter than the sentence encoder with the pre-trained model.
So, the extra constituents in the framework should not be the main cause of the efficiency deterioration. 

\subsubsection{Case Study}
We do the case study in Table \ref{case_study} to show the ability of our model to identify entities in various cases. 
The errors demonstrated in the table are also analyzed.

As can be seen in the first example, our model is capable of recognizing multiple entities presented in long sentences. 
As shown in the table, the proposer is able to produce high-quality proposals. 
And it can be observed from the iterative process of entity suggestions that the regressor regression plays a crucial role in pinpointing entities even if they are relatively short.

Several significant flaws are demonstrated in the second and third cases.
The first problem is that our model has difficulty classifying the entities.
While the model correctly locates the entities in all of its predictions, it makes errors in predicting the type of entities.
We claim that this issue arises as a result of the model's overemphasis on local information.
However, the determination of the entity type usually depends on the global message.

The second problem is that there are some circumstances where the model fails to identify highly overlapped nested entities of relatively short length.
It is understandable because there are considerably more negative data in the training phase than positive ones in this scenario.
Only one proposal represents the genuine entity in most instances where proposals have substantial overlap, while the remaining proposals do not correspond to any entity.
It is partly due to the assignment methodology of the entity proposals and the encoder-only architecture.

\begin{table}
\centering
\caption{Case study. In the left column, the category of the entity is indicated by the label at the bottom right of the right square bracket and the location of the left and right boundary words is shown by the superscript of the square bracket. The iterative process of the entity proposals, as well as the relationship between them and the predicted entities, are shown in the right column.}

\label{case_study}
\resizebox{\linewidth}{!}{
\begin{tabular}{m{0.5\textwidth}c}
\toprule
Sentences with Entities & Predictions \\ \midrule
Treatment of $\color{red}{\text{[}^{2}}$ T cells $\color{red}{\text{]}^{3}_{CEL}}$ with the selective $\color{blue}{\text{[}^{7}}$ PKC $\color{blue}{\text{]}^{7}_{PRO}}$ inhibitor GF109203X abrogates the PMA-induced $\color{blue}{\text{[}^{14}}$ IkB alpha $\color{blue}{\text{]}^{15}_{PRO}}$ phosphorylation/degradation irrespective of activation of Ca(2+)-dependent pathways, but not the phosphorylation and degradation of $\color{blue}{\text{[}^{33}}$ IkB alpha $\color{blue}{\text{]}^{34}_{PRO}}$ induced by TNF-alpha, a $\color{blue}{\text{[}^{40}}$ PKC $\color{blue}{\text{]}^{40}_{PRO}}$-independent stimulus.  
& \begin{tabular}[c]{@{}l@{}}
\Checkmark$ \color{red}{(2,3,CEL)}$ $\gets(2.46,3.06)\gets(2.87,3.01)$ \\ 
\Checkmark$ \color{blue}{(7,7,PRO)}$ $\gets(7.00,6.83)\gets(6.98,7.01)$ \\ 
\Checkmark$ \color{blue}{(14,15,PRO)}$ $\gets(14.41,15.06)\gets(14.76,15.03)$ \\ 
\Checkmark$ \color{blue}{(33,34,PRO)}$ $\gets(33.45,34.04)\gets(33.88,34.01)$ \\ 
\Checkmark$ \color{blue}{(37,37,PRO)}$ $\gets(36.92,36.90)\gets(36.96,37.01)$ \\
\Checkmark$ \color{blue}{(40,40,PRO)}$ $\gets(39.93,39.88)\gets(39.97,40.01)$
\end{tabular}           \\ \hline
Costimulation with $\color{blue}{\text{[}^{2}}$ anti- $\color{blue}{\text{[}^{3}}$ CD28 $\color{blue}{\text{]}^{3}_{PRO}}$ MoAb $\color{blue}{\text{]}^{4}_{PRO}}$ greatly enhanced the proliferative response of $\color{red}{\text{[}^{11}}$ neonatal $\color{red}{\text{[}^{12}}$ T cells $\color{red}{\text{]}^{13}_{CEL}}$ $\color{red}{\text{]}^{13}_{CEL}}$ to levels equivalent to those of $\color{red}{\text{[}^{20}}$ adult $\color{red}{\text{[}^{21}}$ T cells $\color{red}{\text{]}^{22}_{CEL}}$  $\color{red}{\text{]}^{23}_{CEL}}$, whereas  $\color{red}{\text{[}^{25}}$ adult $\color{red}{\text{[}^{26}}$ T cells $\color{red}{\text{]}^{27}_{CEL}}$ $\color{red}{\text{]}^{27}_{CEL}}$ showed only slight increases.
& \begin{tabular}[c]{@{}l@{}}
\Checkmark$ \color{blue}{(3,3,PRO)}$ $\gets(3.12,2.74)\gets(2.98,3.01)$ \\ 
\Checkmark$ \color{blue}{(2,4,PRO)}$ $\gets(3.37,3.96)\gets(3.82,4.02)$ \\ 
\XSolid$ \color{orange}{(11,13,DNA)}$ $\gets(11.95,13.11)\gets(12.61,13.03)$ \\ 
\XSolid$ \color{orange}{(12,13,DNA)}$ $\gets(12.69,13.91)\gets(13.11,14.06)$ \\ 
\XSolid$ \color{orange}{(20,22,DNA)}$ $\gets(21.24,22.15)\gets(21.74,22.02)$ \\
\XSolid$ \color{orange}{(25,27,DNA)}$ $\gets(26.03,27.14)\gets(23.14,24.02)$
\end{tabular}           \\ \hline
Point mutations of either the $\color{orange}{\text{[}^{5}}$ $\color{blue}{\text{[}^{5}}$ PU.1 $\color{blue}{\text{]}^{5}_{PRO}}$ site $\color{orange}{\text{]}^{6}_{DNA}}$ or the $\color{orange}{\text{[}^{9}}$ $\color{blue}{\text{[}^{9}}$ C/EBP $\color{blue}{\text{]}^{9}_{PRO}}$ site $\color{orange}{\text{]}^{10}_{DNA}}$ that abolish the binding of the respective factors result in a significant decrease of $\color{blue}{\text{[}^{25}}$ $\color{blue}{\text{[}^{25}}$ GM-CSF $\color{blue}{\text{]}^{25}_{PRO}}$ receptor  $\color{blue}{\text{]}^{27}_{PRO}}$ alpha promoter activity in $\color{red}{\text{[}^{31}}$ myelomonocytic cells $\color{red}{\text{]}^{32}_{CEL}}$ only.
& \begin{tabular}[c]{@{}l@{}}
\Checkmark$ \color{blue}{(5,5,PRO)}$ $\gets(5.08,4.77)\gets(4.98,5.01)$ \\ 
\XSolid$ \color{red}{(5,6,CEL)}$ $\gets(5.45,6.00)\gets(5.86,6.01)$ \\ 
\XSolid$ \color{red}{(9,10,CEL)}$ $\gets(9.46,9.98)\gets(9.89,10.01)$  \\ 
\Checkmark$ \color{blue}{(25,25,PRO)}$ $\gets(24.91,24.83)\gets(24.97,25.02)$ \\ 
\Checkmark$ \color{blue}{(25,27,PRO)}$ $\gets(26.17,27.02)\gets(26.73,27.03)$ \\
\XSolid$ \color{blue}{(31,32,PRO)}$ $\gets(31.38,32.12)\gets(31.88,32.01)$
\end{tabular}           \\  \bottomrule
\end{tabular}
}
\end{table}

\section{Conclusion and Future Work}
An end-to-end entity detection approach with proposer and regressor is presented in this study.
We employ a proposer that incorporates multi-scale information through the feature pyramid to predict high-quality entity proposals.
It significantly speeds up the training process and boosts performance.
The encoder-only framework is proposed in our work, which introduces significant spatially prior knowledge into the attention mechanism.
It avoids the unfavorable impact of random initialization of query vectors.
In order to increase prediction accuracy, we explore iterations over span locations and category logarithms in the joint model.
We model the entity proposal and prediction target assignment issue as a Linear Assignment Problem and compute the bipartite loss during the training phase.
The model is able to predict the entities in a single run.
The experiments on datasets of different characteristics demonstrate the nature of our model and the effectiveness of our approach.
Our work reveals the potential to integrate spatial priors into NLP research.
We expect the findings will contribute to a better understanding of the set prediction network and iterative refinement.
We will go deep into the relevance of NLP and CV tasks in the future.

\subsubsection*{Acknowledgements} 
This work was supported by the National Natural Science Foundation of China under Grant 62072211, Grant 51939003, and Grant U20A20285.

\subsubsection*{Code Availability} 
The code is available at \href{https://github.com/Rosenberg37/EntityDetection}{https://github.com/Rosenberg37/EntityDetection}.

\section*{Declarations}
\subsubsection*{Conflicts of interest} 
The authors declare that they have no known competing financial interests or personal relationships that could have appeared to influence the work reported in this paper.

\bibliographystyle{unsrt}  
\bibliography{references}  

\end{document}